\documentclass[11pt,a4paper]{article}
\usepackage[hyperref]{emnlp-ijcnlp-2019}
\usepackage{times}
\usepackage{latexsym}
\usepackage{helvet}  
\usepackage{courier}  
\usepackage{url}  
\usepackage{graphicx}  
\usepackage{booktabs} 
\usepackage{tabularx}
\usepackage{multirow}
\pdfmapfile{+txfonts.map}
\usepackage{caption}
\usepackage{subcaption}
\usepackage{changepage}

\usepackage{amsmath,amssymb,amsfonts}
\usepackage{algorithmic}
\usepackage{textcomp}
\usepackage{xcolor}
\aclfinalcopy 

\title{Towards Controllable and Personalized Review Generation}

\author{Pan Li \\
  New York University \\ Stern School of Business \\
  44 West 4th Street, NY, NY, 10012 \\
  {\tt pli2@stern.nyu.edu} \\\And
  Alexander Tuzhilin \\
  New York University \\ Stern School of Business \\
  44 West 4th Street, NY, NY, 10012 \\
  {\tt atuzhili@stern.nyu.edu} \\}

\date{}

\begin{document}
\maketitle
\begin{abstract}
In this paper, we propose a novel model \textit{RevGAN} that automatically generates \textit{controllable} and \textit{personalized} user reviews based on the arbitrarily given sentimental and stylistic information. RevGAN utilizes the combination of three novel components, including self-attentive recursive autoencoders, conditional discriminators, and personalized decoders. We test its performance on the several real-world datasets, where our model significantly outperforms state-of-the-art generation models in terms of sentence quality, coherence, personalization and human evaluations. We also empirically show that the generated reviews could not be easily distinguished from the organically produced reviews and that they follow the same statistical linguistics laws.
\end{abstract}

\section{Introduction}
With the ever increasing interests in user-generated reviews on online marketplace websites, such as Amazon, Yelp and TripAdvisor, it is necessary to provide a range of tools that would encourage users to provide feedback in a more efficient and effective manner, as only a small fraction of users really take time to write their own reviews \cite{chen2008online}. Automatic review generation, for example, takes the product information and user behavior as input and generates user reviews following the arbitrarily given users' sentiment designation and writing style personalized towards the specific product and user.

Researchers have proposed various types of product review generation methods \cite{yao2017automated,dong2017learning,lipton2015generative,radford2017learning} and achieved great performance. However, they did not consider the inner hierarchical word-sentence-paragraph structure within user reviews, thus making their generation results significantly limited in length and coherence. \cite{li2015hierarchical,zang2017towards} did include the hierarchical connection in their review generation model, but they did not address the problem of \textit{controllable} and \textit{personalized} review generation targeted at the specific product and user, which is essential for the usefulness of generated reviews. Most importantly, all the aforementioned generative models did not include \textit{production descriptions} in the generation process, thus their generation results lack credibility and diversity.

To address these problems, we propose a novel model \textit{RevGAN} that automatically generates high-quality user reviews given the information of product descriptions, sentiment labels and users' historical reviews. The proposed RevGAN model follows a three-staged process: In Stage 1, we propose to use \textit{Self-Attentive Recursive Autoencoder} for mapping the discrete user reviews and product descriptions into continuous embeddings for the advantage of capturing the `'essence'' of textual information and the convenience for subsequent optimization processes. In Stage 2, we utilize a novel \textit{Conditional Discriminator} structure to control the sentiment of generated reviews by conditioning sentiment on the discriminator and forced the generator to adapt its generation policy correspondingly. Finally in Stage 3, to improve the personalization of generated reviews, we use a new \textit{Personalized Decoder} method to decode the generated review embeddings according to users' writing styles extracted from their history corpus.

We conduct extensive experiments using multiple real-world datasets and show that the proposed RevGAN model significantly and consistently outperforms state-of-the-art baseline models and lead to the automated generation of reviews that are indeed very similar in style and content to the set of original reviews.

In general, this paper makes the following contributions:

a. We propose a novel \textit{RevGAN} model that automatically generates \textit{controllable} and \textit{personalized} reviews from product information, a set of user reviews and their writing styles. Especially. we propose three novel components of the generative framework: \textit{Self-Attentive Recursive Autoenocder} that captures the hierarchical structure and latent semantic meanings of user reviews, \textit{Conditional Discriminator} that generates controllable user reviews by conditioning the sentimental information on the discriminator to improve the generation performance in terms of sentence quality and context accuracy, and \textit{Personalized Decoder} that takes the personalized writing style into account by concatenating the users' vocabulary preference onto the decoder to improve the personalization and credibility of the generated results, which is validated by the empirical human evaluation. 

b. We empirically demonstrate that our proposed RevGAN model achieves state-of-the-art review generation performance, statistically and empirically outperforming several important benchmarks on multiple datasets. We also empirically show that the reviews generated by our method are very similar to the organically generated reviews and that the linguistic features of generated reviews follow the same statistical linguistics laws as reviews organically produced by the users.
\section{Related Work}
In this section, we briefly summarize the related work following two aspects covering previous work on automated review generation and GAN for NLG. We point out the connection and difference between our proposed model and prior literature, which leads to significant improvements of review generation performance.
 
\subsection{Automated Review Generation}
Researchers have been utilizing multiple versions of Seq2Seq\cite{sutskever2014sequence} framework to generate online product reviews of good quality, including Aspect-Aware Representations \cite{ni2018personalized}, Gated Contexts to Sequences\cite{tang2016context}, RNN\cite{yao2017automated}, Aspect-Sentiment Score\cite{zang2017towards}, Generative Concatenative Nets\cite{lipton2015generative} and Sentiment Units\cite{radford2017learning}. In particular, \cite{li2015hierarchical} proposed a two-stage LSTM neural network to construct a hierarchical autoencoder for long-text representation and generation. 

However, these review generative models include neither product information nor users' writing styles into the generation process, thus making the generated reviews less persuasive. Also, they are limited in length and lack of coherence for neglecting the hierarchical connections within sentences, which are very important elements towards the helpfulness of a specific review\cite{mudambi2010research}. Our RevGAN model, on the other hand, utilizes the combination of self-attentive hierarchical autoencoder and conditional discriminator for improved and controllable review generation, while we concatenate the contextual labels and users' history corpus into the personalized decoder at the same time. Experimental results support that our proposed model indeed achieves significantly better generation results compared to the prior literature.

\subsection{GAN for NLG}
GAN\cite{goodfellow2014generative} has become a powerful method for reconstruction and generation in real data space, which leaves great potential to be used for natural language generation purposes. Various methods have been proposed to get over the major problem of the discontinuity of textual information, including SeqGAN\cite{yu2017seqgan}, TextGAN\cite{zhang2017adversarial}, RankGAN\cite{lin2017adversarial} and LeakGAN\cite{guo2017long}. However, regarding long-text generation tasks, for all the models above, the computation complexity might be too high, thus failing to provide satisfying results. Nor do these models take contextual and personalized information into consideration for controllable generation. Conditional GAN\cite{mirza2014conditional,hu2017toward,dong2017learning} concatenates the supervised labels into the input of generator and is able to control the generation of simple sentences. However, considering the high dimension of latent embedding vectors, concatenating significantly lower dimensional supervised information into the input might not be strong enough to force the generator to update towards the designated direction. 

Therefore, to address these problems, we propose a novel conditional discriminator model which conditions the sentiment label on the discriminator to artificially change how the discriminator works, and force it backpropagate the loss function that could make the generator to learn what the user really want. Experimental results reported in Section 5 show that we outperform all other GAN-based-NLG models in the generation performance, and outperform Conditional GAN in terms of sentiment accuracy as well.


\section{Models}
In this section, we introduce the proposed \textit{RevGAN} model for review generation that includes three novel components: Self-Attentive Recursive AutoEncoder, Conditional Discriminator and Personalized Decoder. We first use Self-Attentive Recursive AutoEncoder to map the discontinuous user reviews and product descriptions into a latent continuous space, and utilize a novel version of cGAN to generate review embeddings for subsequent personalized decoding and review generation. Experimental results show that the combination of all three novel components achieves state-of-the-art review generation performance.

\begin{figure}[h]
\centering
\includegraphics[width= 0.4\textwidth]{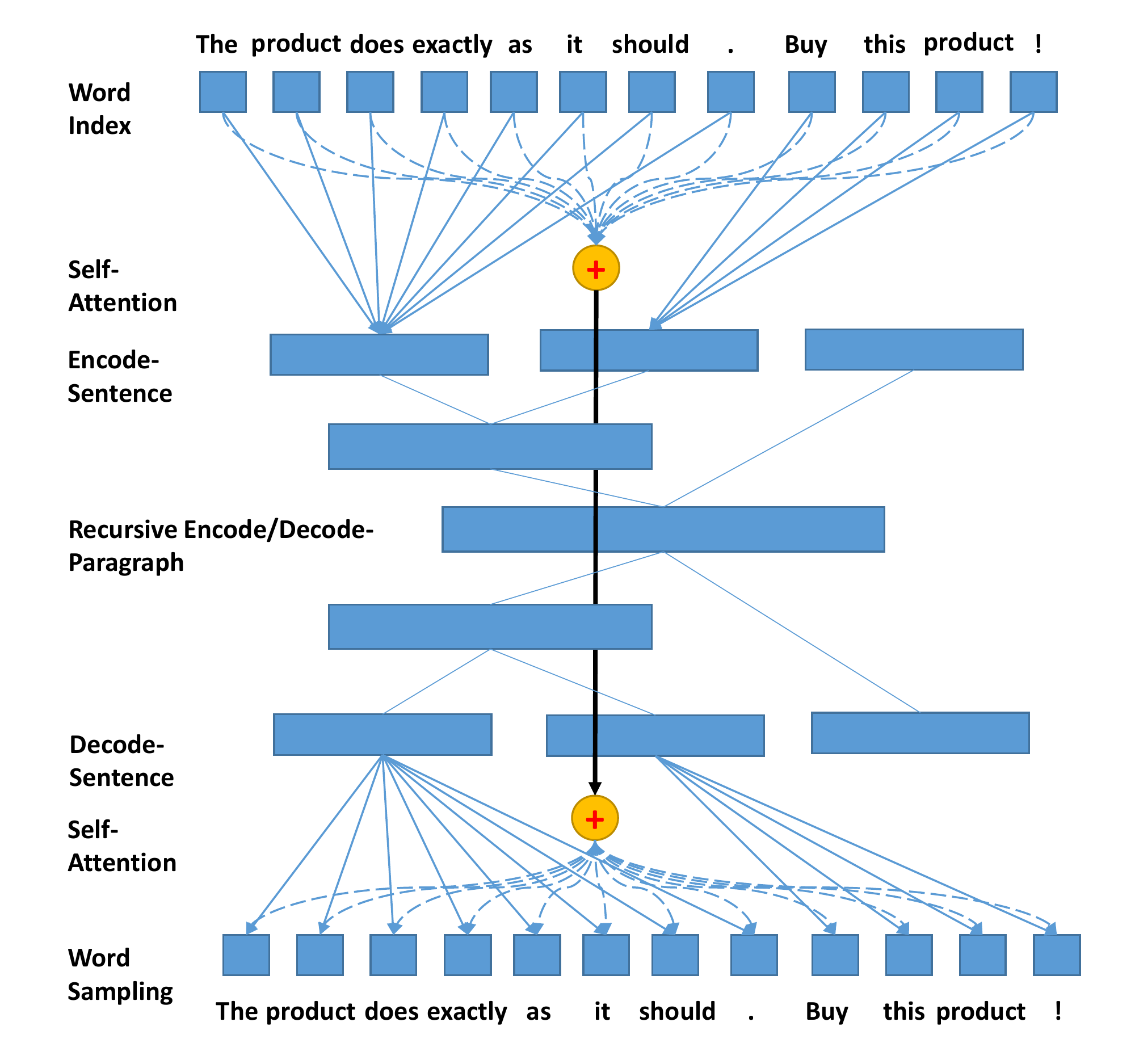}
\caption{A Brief Overview of Self-Attentive Recursive AutoEncoder}
\label{process}
\end{figure}

\subsection{Self-Attentive Recursive AutoEncoder}
The proposed Self-Attentive Recursive AutoEncoder is illustrated in Figure \ref{process}. We implement a bidirectional Gated Recurrent Unit (GRU) \cite{cho2014learning} neural network as the encoder and the decoder in the model respectively. Compared with classical models like RNN or LSTM, GRU is computationally more efficient and better captures latent semantic meanings. We split each user review or product description into single sentences, and then map each sentence to their corresponding word indexes in the pre-defined dictionary. The index sequences constitute the input of our proposed model.
 
We use $R$ to represent a certain review consisting of $N^{R}$ sentences as $R=\{s^{1},s^{2},\cdots,s^{N_{R}}\}$. Each sentence $s$ consists of $N^{s}$ words as $s=\{w^{1},w^{2},\cdots,w^{N_{s}}\}$, where $w_{i}$ represents the index of certain word in the vocabulary with size $V$. We denote $[W^{z},W^{r},U^{z},U^{r}]$ as the weight matrices for update gates and reset gates, $z_{t}, r_{t}$ as the status for the update gate and reset gate, and $x_{t}, h_{t}$ as the input and output vector at time $t$ respectively. GRU learns the hidden representations using the following equations, while the hidden state $h_{t}$ at the end of the sequence constitutes latent sentence embeddings. 
$$z_t = \sigma_g(W_{z} x_t + U_{z} h_{t-1} + b_z) $$
$$r_t = \sigma_g(W_{r} x_t + U_{r} h_{t-1} + b_r) $$
$$h_t =  (1-z_t)h_{t-1} + z_t\sigma_h(W_{h} x_t + U_{h} (r_t h_{t-1}) + b_h) $$

\begin{figure}[h]
\centering
\includegraphics[width= 0.2\textwidth]{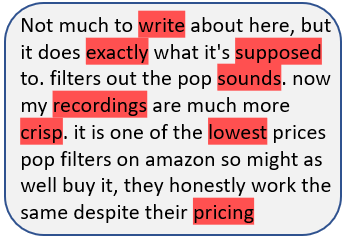}
\caption{Visualization of Self-Attentive Mechanism}
\label{attention}
\end{figure}

Besides, to capture the relative position representations. we incorporate self-attentive mechanism \cite{shaw2018self} during encoding process. Typically,  each output element $h_{i}$ is computed as weighted sum of a linearly transformed input elements $$h_{i}=\sum_{j=1}^{n}\alpha_{ij}(x_{j}W^{V})$$ Each weight coefficient $\alpha_{ij}$ is computed using a softmax function $$\alpha_{ij}=\frac{exp e_{ij}}{\sum_{k=1}^{n}exp e_{ik}}$$ And $e_{ij}$ is computed using a compatibility function that compares two input elements correspondingly. We visualize the self-attention mechanism with an example in Figure \ref{attention}.

After getting the sentence-level embeddings within each review, we merge those sentence embeddings in a recursive way to obtain paragraph embeddings via a binary tree structure. We denote the embedding for sentence s as $e_{s}$, and during the encoding process the first parent node $y_{2}$ is computed from the first two children node $(e_{1},e_{2})$ by the standard neural network layer:
\[y_{2}=f(W_{e}(e_{1}:e_{2})+b)\]
where $(e_{1}:e_{2})$ is the natural concatenation of these two embedding vectors, $W_{e}$ is the weight matrix with twice the size of the embedding vectors. The second parent node $y_{3}$ will be computed from the concatenation of the first parent node $y_{2}$ and the following embedding vector $e_{3}$:
\[y_{3}=f(W_{e}(y_{2}:e_{3})+b)\]
We obtain the representation of the rest nodes similarly. The embedding for root node constitutes the entire review embeddings. The training process of $W_{e}$ follows a standard MLP network, where we optimize the reconstruction loss over every layer of the binary tree.

To unfold the recursive autoencoder, we'll start from the root node $y_{N^{R}}$ of the binary tree. By utilizing another MLP network, we expand the paragraph embedding vector to two vectors: the leaf node and the lower level parent node, where the parent node would go through the same procedure until all of the leaf nodes in the binary tree are deciphered.
\[(y_{N^{R}-1}:e_{N^{R}})=f'(W'_{e}y_{N^{R}}+b)\]
Finally, we assemble the paragraph from the bottom leaf node to the top one such that $R=\{e_{1},e_{2},\cdots,e_{N^{R}}\}$ to complete the review reconstruction process.

\subsection{Conditional Discriminator}
To generate meaningful reviews from the product information, we construct a cGAN model to transfer product descriptions into user reviews given specific sentiment labels. However, unlike the traditional conditional GAN methods \cite{mirza2014conditional,hu2017toward}, we do not concatenate the sentiment label directly into latent codes; considering the relatively high dimensions of latent embeddings, concatenating the sentiment scalar into the input might not be powerful enough to force the generator to update itself to match with the designated sentiment. Thus, we condition the sentiment labels on the discriminator to artificially change the rules that the discriminator works, and force it backpropagate loss functions that update generator policy correspondingly. The generated reviews and original reviews with the opposite sentiment are judged as negative examples, while only the original reviews that matches with the given sentiment are judged as positive examples by the conditional discriminator, as we propose the novel conditional discriminator D (for positive sentiment):
$$
D(x|c)=\left\{
\begin{aligned}
1,& \text{organic positive reviews} \\
0,& \text{others} \\
\end{aligned}
\right.
$$

For generator, the model is optimized to minimize the reconstruction error between generated reviews and original reviews, $L(\theta^{G})=KL(p_{real}(x)||p_G(x;\theta))$
where KL stands for Kullback-Leibler divergence. The loss function of discriminator D is: $L(\theta^{D},\theta^{G})=-\frac{1}{2}E_{x\sim data}logD(x)-\frac{1}{2}E_{x}log(1-D(G(z)))$ Under ideal circumstances when generator and discriminator both reach their equilibrium, we could get that $D(x)=D(G(x))=\frac{1}{2}$ and that the generated reviews are indeed indistinguishable from the original reviews from the discriminator point of view.
 
The core idea lies in that, by artificially forcing the discriminator to take certain type of reviews as real samples, generator should learn the conditioned information and then transform the generated data distribution. This unique structure of GANs makes possible the controllable review generation process, and experimental results support the strength of our model over classical cGAN models.

\subsection{Personalized Decoder}
To personalize the generation process, apart from the conditioning of sentiment labels, we also take the users' specific writing styles into account. We provide the definition of writing style according to \cite{zheng2006framework} :\\
\textbf{Writing Style} refers to the user's distinctive vocabulary choices and style of expression in his review creations.\\
Assuming that the historical reviews written by user i contain $[R_{i1},R_{i2},\cdots,R_{iN_{i}}]$, we calculate the usage frequency of each word from the corpus, which is denoted as a V-dimensional writing style vector $W_{style}$. The intuition is that, during the decoding process, instead of generating each word right from the calculated word distribution via GRU network, we concatenate the writing style vector onto the distribution and sample the generated word afterwards, which would be determined by both the writing style vector $W_{style}$ and the distribution vector $W$:
\[W'=W*W_{style}\]
Note that, to deal with the cold start problem when the user has no historical reviews, we could simply set the writing style vector as identity matrix $W_{style}=I$ and generate the reviews under normal settings. Experimental results show that the involvement of personalized information (sentiment information and writing style) indeed improve the generation results and the helpfulness score from the empirical study as well.
\section{Experimental Results}
\subsection{Dataset}
To empirically validate our proposed model, we implemented RevGAN on three subsets of the Amazon Review Dataset\cite{he2016ups,mcauley2015image}\footnote{http://jmcauley.ucsd.edu/data/amazon/}, namely Musical Instrument, Automotive and Patio which include 44,006 reviews written by 3,697 users on 6.039 items.

\subsection{Experiment Settings}
The self-attentive recursive autoencoder is implemented by bidirectional GRUs with embedding dimension 300. GRU parameters and word embeddings are initialized from a uniform distribution between [-0.1,0.1]. The initial learning rate is 1e-3, which will be halved every 50 epochs until convergence. Batch size is set to 128 (128 sentences across review documents) for batch normalization\cite{ioffe2015batch}. Sentences would be padded to the maximum length within each batch. Gradient clipping\cite{gulrajani2017improved} is adopted by scaling the gradients when the norm exceeds the threshold 1. For the recursive structure, the parameter settings are the same with sentence-level autoencoder only except that the size of the weight matrix is $600\times300$. The beam size for beam searching\cite{wiseman2016sequence} would be fixed as 3. To validate the emotion label for each review, we implemented the state-of-the-art sentiment classifier VADER\cite{gilbert2014vader}\footnote{https://github.com/cjhutto/vaderSentiment} to label the sentiment score for each review. The baseline SeqGAN, RankGAN and LeakGAN models are implemented through the Texygen\cite{zhu2018texygen}\footnote{https://github.com/geek-ai/Texygen} toolkit. The generator and the conditional discriminator of GANs are both set as Multilayer Perceptron\cite{rumelhart1985learning} (MLP) with 300 hidden layers. Their parameters are initialized from the normal distribution N(0,0.02). The learning rates for generator and conditional discriminator are fixed at 5e-5 and 1e-5 respectively. During each epoch, generator G would iterate 5 times while discriminator D would only iterate 1 time. The model updates 30,000 times in total. We implemented our model on a Tesla K80 GPU within PyTorch\footnote{https://pytorch.org/} environment, where the whole training takes about 12 hours.

\subsection{Evaluation Metrics}
To demonstrate that our purposed model indeed achieves the state-of-the-art review generation performance, we implement various evaluation metrics, including distribution-based Log-Likelihood and Perplexity, coherence-based Word Mover Distance (WMD) \cite{kusner2015word}, ngram-based BLEU \cite{papineni2002bleu} and ROUGE \cite{lin2004rouge}, contextual label accuracy and human evaluation to measure the performance of review generation. Specifically, following the same metric as \cite{dong2017learning}, we use sentiment accuracy, the ratio of the reviews whose sentiment matches with the given label, as an important indication of the personalization ability of the generator. The higher sentiment accuracy is, the better it could provide supervised generated results. Besides, we conduct the human evaluation to assess the quality and helpfulness of generated results by randomly selecting the same number of reviews from the original dataset, the generation of RevGAN and the generation of other baseline models and asking the participants to analyze which ones are generated by the machine and which ones are really created by humans, where significance test shows that our generated reviews are indeed indistinguishable from the original data.

\subsection{Baseline Models}
To demonstrate that our purposed model indeed achieves the state-of-the-art review generation performance, we compare our model across various evaluation metrics with several important benchmarks, including charRNN \cite{yao2017automated}, MLE \cite{bahl1990maximum}, SeqGAN \cite{yu2017seqgan}, LeakGAN \cite{guo2017long}, RankGAN \cite{lin2017adversarial} and Attr2Seq \cite{dong2017learning}. Besides, to verify the effectiveness of combining three novel components into the RevGAN model, we also compare the performance between RevGAN+CD (Conditional Discriminator), RevGAN+CD+SA (Self-Attentive Autoencoder) and RevGAN+CD+SA+PD (Personalized Decoder). The results show that our model indeed outperforms all the selected benchmark models significantly and consistently.

\subsection{Significance Testing}
We conduct significance tests to identify whether the difference between two review generation algorithms could indicate a difference in true system quality. Typically, following\cite{koehn2004statistical}, we use bootstrap re-sampling methods to get the asymptotic standard error of the estimated value of the evaluation metrics. Then the paired two-sample t-test could be used to test the significance whether their population means differ statistically.

In terms of the indistinguishableness of the generated results, we conduct a chi-square test for independence to test whether there is a significant association between the human assessment and the actual value. As the results of statistical test are insignificant, we could then claim that our generated reviews are indistinguishable from original ones in the sense that human can't separate them apart.

\begin{table*}
\small
\centering
\begin{tabular}{cccccc} \hline
Models & Log-Likelihood & WMD & PPL & BLEU-4(\%) & ROUGE-L(\%) \\ \hline
SeqGAN & -86699 & 1.869 & 22.60 & 15.06 & 38.30 \\
LeakGAN & -108581 & 2.324 & 24.09 & 14.98 & 37.73 \\
RankGAN & -73309 & 1.862 & 22.45 & 14.92 & 37.72 \\
charRNN & -100430 &1.976 & 22.07 & 11.46 & 33.60 \\
MLE & -54338 & 2.106 & 17.15 & 9.62 & 31.89 \\
Attr2Seq & -56298 & 2.077 & 21.00 & 11.48 & 32.22 \\ \hline
RevGAN+CD & -80386 & 2.097 & 19.71 & 21.32 & 39.47 \\
RevGAN+CD+SA & -51549 & 2.030 & 17.45 & 24.44 & 41.32 \\
RevGAN+CD+SA+PD & \textbf{-34305**} & \textbf{1.762**} & \textbf{17.00*} & \textbf{27.16**} & \textbf{44.63**} \\ \hline
\end{tabular}
\caption{Comparison of Experimental Results on Amazon Review Dataset (** stands for significance under 99\% confidence, * stands for 95\% confidence)}
\label{basicresults}
\end{table*}

\subsection{Evaluation of review generation}
To illustrate the superiority and generalizability of our RevGAN model, we implement our model on three different domains of the Amazon Review Dataset including musical instruments, automotive and patio products. The summary of our experiment results is reported in Table \ref{basicresults}, from which we could clearly observe that, compared with the baseline text-generation models, our proposed RevGAN model performs significantly better in sentence quality and coherence performance. On average, we could witness a 5\% increase in Word Mover Distance (WMD), 80\% improvement in BLEU and 10\% rising in ROUGE.  Besides, the comparison between different variations of RevGAN model verifies that indeed the combination of all three novel components gives the best generation performance. By deploying bootstrap re-sampling techniques introduced in the previous section, we conduct hypothesis tests where all the tests confirm the significant improvement of our RevGAN model. In that sense, we claim that our model achieved the state-of-the-art results on review generation. We also showcase some generated reviews at the end of this section.

\subsection{Evaluation of controllable generation}
In this part, we evaluate the controllable generation performance of our purposed model by pre-setting the contextual labels. We fixed the sentiment label as 'positive' and 'negative' respectively conditioned on the discriminator, and then evaluate the sentiment accuracy of the generated reviews. The results are reported in Table \ref{sentimenttransfer}, where our model beats the state-of-the-art algorithm Attr2Seq\cite{dong2017learning} and the classical model Conditional GAN where we condition the same sentiment on these two models as well.


\begin{table}[h]
\small
\centering
\begin{tabular}{cccc} \toprule
	Metrics & RevGAN+CD & Attr2Seq & cGAN\\ \midrule
	Accuracy &  \textbf{0.842*} & 0.762 & 0.665\\ \bottomrule
\end{tabular}
\newline
\caption{Controllable Generation Accuracy}
\label{sentimenttransfer}
\end{table}


\subsection{Evaluation of Personalized Generation}
Besides the statistical and semantical metrics, we also design an empirical study to test the personalized performance of our generated reviews. We randomly select 15 reviews to include in each questionnaire, 5 from the original dataset, 5 from RevGAN generated results with personalization and 5 from RevGAN generated results without personalization, and ask participants to analyze which ones are generated by the machine and which ones are really created by humans. Besides, they are also asked to assess the helpfulness of each review by choosing the helpfulness score scale 1-5 for each review. We sent out 100 questionnaires in total, and get 36 responses, the confusion matrix of which is reported in Table \ref{confusionmatrix}. To test that whether the RevGAN generated reviews are indeed statistically indistinguishable from original ones, we run chi-test for significance testing:
\[\chi^{2}=0.012 < \chi_{0.95}^{1}\]
which shows that, under 95\% confident interval, we could claim that there's no statistical difference between our machine-generated reviews and those actually written by humans. 

Besides, the results indicate that our generated reviews have no statistical difference in terms of the helpfulness scores from those written by consumers towards certain products, with average helpfulness scores 3.10 and 3.03 for machine-generated and real-world reviews respectively. Thus, based on the t-test, we accept the hypothesis that there's no statistical difference between those two groups in terms of helpfulness as well.
\[t= \frac{\bar{X_{1}}-\bar{X_{2}}}{s_{p}\sqrt{\frac{2}{n}}} = 0.77 < t_{0.95}\]
And finally, we would conduct t-test over the performance of personalized and non-personalized generated results with helpfulness score 3.10 and 2.91, which indicates the significant improvement in helpfulness by the involvement of users' writing style.
\[t= \frac{\bar{X_{1}}-\bar{X_{2}}}{s_{p}\sqrt{\frac{2}{n}}} = 2.09 > t_{0.95}\]

\begin{table}[h]
\small
\centering
\begin{tabular}{|c|c|c|}
\hline
 &\multicolumn{2}{c|}{Actual Value}\\
\cline{2-3}
Empirical Test & Human-Written & RevGAN+PD \\ \hline
Human & $119$ & $61$ \\ \hline
Machine & $118$ & $62$ \\ \hline
\end{tabular}
\begin{tabular}{|c|c|c|}
\hline
 &\multicolumn{2}{c|}{Actual Value}\\
\cline{2-3}
Empirical Test & Human-Written & RevGAN \\ \hline
Human & $119$ & $61$ \\ \hline
Machine & $102$ & $78$ \\ \hline
\end{tabular}
\newline
\caption{Confusion Matrix of Empirical Test}
\label{confusionmatrix}
\end{table}

\subsection{Showcase}
\begin{table*}[!]
\small
\begin{tabular}{|p{0.09\textwidth}|p{0.09\textwidth}|p{0.7\textwidth}|}
\hline
Domain & Sentiment & Generated Reviews  \\ \hline
Musical & Positive & These chords got me to play my guitar better in less than one day. An excellent overdrive and an incredible value. I'll use them all the time.\\ \hline
Musical & Negative & These pedals are not budget friendly. If you are looking for classic rock sounds, you won't love these expensive hardware.\\ \hline
Automotive & Positive & I bought two sets of seat covers and this roll kit. Both fit well and look good. They were much easier to slide over the leather seats. \\ \hline
Automotive & Positive & These seat covers look good and seem to be made of a good quality material. For the price, these are a great buy. \\ \hline
Patio & Negative & It is not recommended. The cover is a little tight and hard to open and close.\\ \hline
Patio & Positive & These traps have caught more mice than ever give. You only need a little peanut butter for the bait and tomcat would caught so many mice in one night. Will order again if needed .\\ \hline
\end{tabular}
\newline
\caption{Selected Personalized Generation Examples}
\label{personalized}
\end{table*}

\begin{table*}[h]
\small
\begin{tabular}{|p{0.50\textwidth}|p{0.19\textwidth}|p{0.19\textwidth}|}
\hline
User History Reviews & RevGAN & RevGAN+PD \\ \hline
1.They play well and hold up well(never had one break) They are my second favorite pick. But this pick is \textbf{better} in some songs than my favorite. 

2.I ordered 5 different kind of picks and these were my favorite picks. They have a very \textbf{comfortable feel} and great sound!
& The guitar has always made a quality and you would really love that life from it! & The guitar has always made \textbf{better} quality and you would really love that \textbf{comfortable feel} from it! \\ \hline
\end{tabular}
\newline
\caption{Personalized Review Generation Examples}
\label{person}
\end{table*}

We present several showcases of our generated results with different contextual labels and domains as shown in Table \ref{personalized}. Additionally, we showcase the modification process in Table \ref{person}, where the personalized generated reviews tend to use more words from the user's history corpus.

\begin{figure}[h]
\centering
\includegraphics[width=0.4\textwidth]{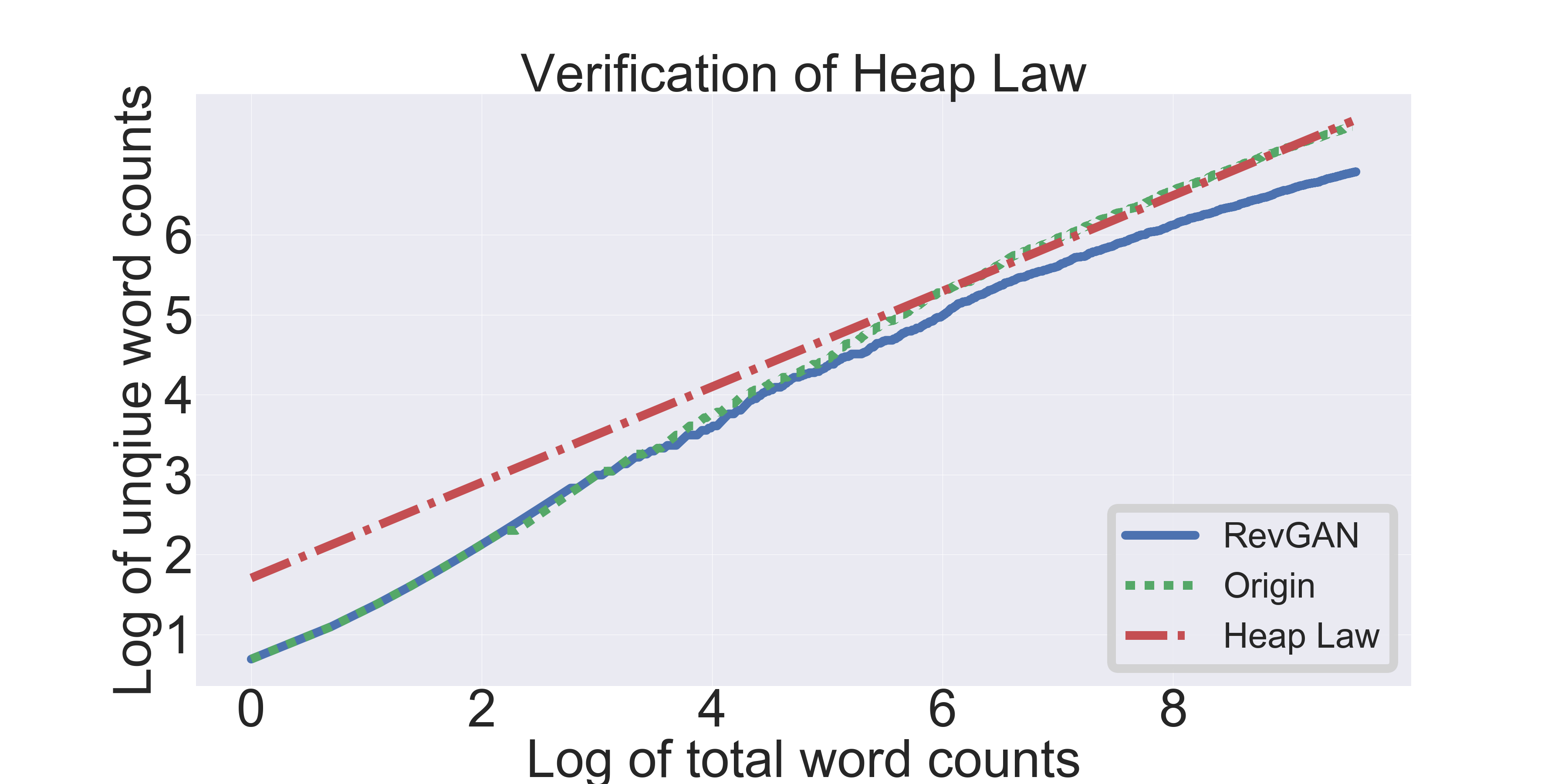}
\caption{Verification of Heap's Law}
\label{verfication1}
\end{figure}

\begin{figure}[h]
\centering
\includegraphics[width=0.4\textwidth]{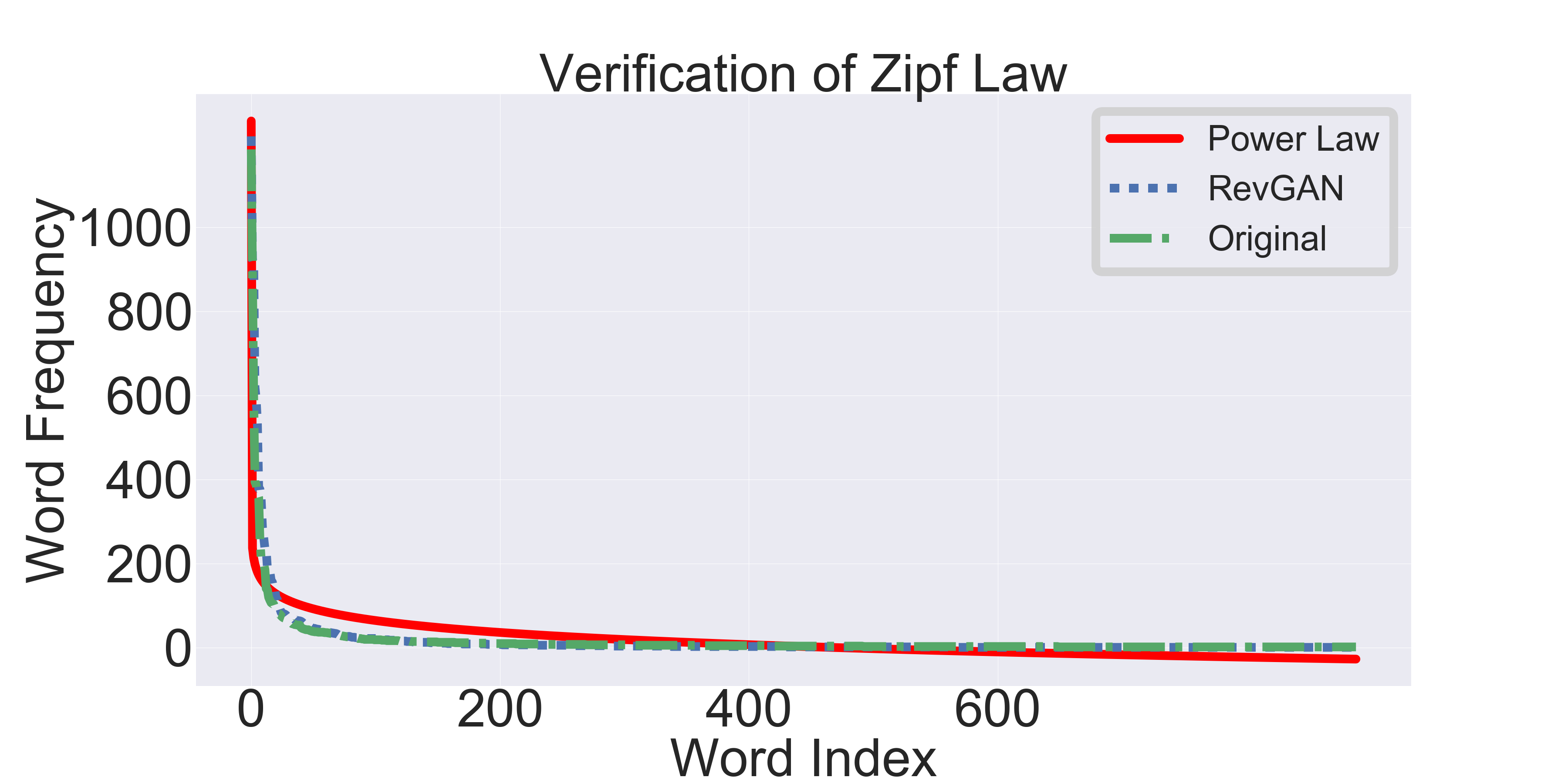}
\caption{Verification of Zipf's Law}
\label{verfication2}
\end{figure}

Besides, we check if reviews generated by RevGAN would have the same linguistic features by testing two major statistical laws of linguistics\cite{altmann2016statistical}: Zipf Law\cite{zipf1935psycho} and Heap Law\cite{herdan1964quantitative}. The former states that if words are ranked according to their frequency of appearance $r=1,2,\cdots,V$, the frequency f(r) of the r-th word scales with the rank as $f(r)=\frac{\beta_{Z}}{r^{\alpha_{Z}}}$, while the latter states that the number of different words V scales with database size $N$ measured in the total number of words as $V \sim N^{\alpha_{H}}$. As shown in Figure \ref{verfication1} and \ref{verfication2}, both the original reviews and the generated ones satisfy those two linguistic laws. 

\section{Conclusion}
In this paper, we proposed \textit{RevGAN} that automatically generates personalized product reviews from product embeddings as opposed to labels, which could output results targeting on specific products and users. To do this, we incorporate three novel components: self-attentive recursive autoencoder, conditional discriminator and personalized decoder.  Experimental results show that RevGAN performs significantly better than other baseline models and that  our generated reviews are very similar to organically generated user reviews, as shown in Section 5.2 and Table \ref{confusionmatrix}.

As a part of the future work, we would like to improve the review generation process in a way that  could receive several key words from users as input and generate reviews based on these prior information. Another direction of the future research, however, lies in developing novel methods that distinguish the type of reviews described in the paper and organic reviews.

\bibliographystyle{acl_natbib}
\bibliography{sigproc}

\end{document}